\pdfoutput=1

\documentclass[11pt]{article}

\usepackage[final]{acl}

\usepackage{times}
\usepackage{latexsym}
\usepackage{multirow}
\usepackage{enumitem}

\usepackage[T1]{fontenc}

\usepackage[utf8]{inputenc}
\usepackage{tikz}

\usepackage{microtype}

\usepackage{inconsolata}

\usepackage{graphicx}
\usepackage{xspace,mfirstuc,tabulary}
\usepackage{booktabs}
\usepackage{todonotes}
\usepackage{tcolorbox}
\usepackage{makecell}
\usepackage{amssymb}
\usepackage{caption}
\usepackage{array}
\usepackage{colortbl} %
\usepackage{xcolor}
\usepackage{ulem}
\usepackage{arydshln}
\usepackage{amsmath}
\usepackage{pifont}

\definecolor{lightyellow}{RGB}{255, 255, 200}
\definecolor{red-arrow}{RGB}{224, 105, 50}
\definecolor{blue-arrow}{RGB}{13, 202, 242}

\newcommand*\circled[1]{\tikz[baseline=(char.base)]{
            \node[shape=circle,draw,inner sep=.6pt] (char) {#1};}}

\newcommand{\cmark}{\ding{51}}
\newcommand{\xmark}{\ding{55}}

\title{Limited-Resource Adapters Are Regularizers, Not Linguists}

\author{Marcell Fekete$^{1}$ \ \ Nathaniel R. Robinson$^{2}$ \ \  Ernests Lavrinovics$^{1}$ \ \  \\  {\bf E. Djeride Jean-Baptiste$^{3}$ \ \ Raj Dabre$^{4}$ \ \ Johannes Bjerva$^1$ \ \  Heather Lent$^{1}$ } \\
         $^1$Aalborg University, Denmark;
          $^2$Johns Hopkins University, USA; \\
          $^3$University of Ottawa, Canada; $^4$IIT Madras, India \\
          \texttt{\{mrfe, hcle\}@cs.aau.dk} \\ 
}

\begin{document}
\maketitle
\begin{abstract}

Cross-lingual transfer from related high-resource languages is a well-established strategy to enhance low-resource language technologies. 
Prior work has shown that adapters show promise for, e.g., improving low-resource machine translation (MT). In this work, we investigate an adapter souping method combined with cross-attention fine-tuning of a pre-trained MT model to leverage language transfer for three low-resource Creole languages, which exhibit relatedness to different language groups across distinct linguistic dimensions. 
Our approach improves performance substantially over baselines. 
However, we find that linguistic relatedness---or even a lack thereof---does not covary meaningfully with adapter performance. 
Surprisingly, our cross-attention fine-tuning approach appears equally effective with randomly initialized adapters, implying that the benefit of adapters in this setting lies in parameter regularization, and not in meaningful information transfer. 
We provide analysis supporting this regularization hypothesis. 
Our findings underscore the reality that neural language processing involves many success factors, and that not all neural methods leverage linguistic knowledge in intuitive ways. 

\end{abstract}

\section{Introduction and Background}\label{sec:intro}

\begin{figure}
    \centering
    \includegraphics[width=\columnwidth]{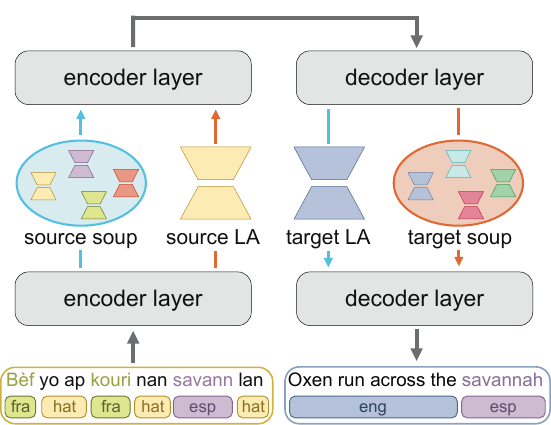}
    \caption{
    Overview of the MT transfer experiments between English and Creoles. The
    \mbox{\textcolor{red-arrow}{$\boldsymbol{\uparrow}$} path} shows the use of source language adapter in the encoder and the `souped' (weight-averaged) target and transfer language adapters in the decoder layers. The
    \textcolor{blue-arrow}{$\boldsymbol{\uparrow}$} path shows the opposite direction.
    }
    \label{fig:enter-label}
\end{figure}

For a majority of the world's languages, data scarcity remains a barrier for reliable machine translation (MT).
Cross-lingual transfer from high-resource to low-resource languages has been shown to help mitigate this \cite{zoph-etal-2016-transfer, tars-etal-2021-extremely}, but has still not closed the gap between high- and low-resource scenarios. 
In this work, we seek to exploit methods from parameter-efficient fine-tuning (PEFT) to further improve cross-lingual transfer for a Creoles, an under-served group of low-resource languages.

PEFT methods such as bottleneck adapters \citep{houlsby_parameter-efficient_2019,pfeiffer-etal-2020-mad}, LoRA \citep{hu_lora_2021}, and (IA)$^3$ \citep{liu_few-shot_2022} are widely used to adapt language models to low-resource languages \cite{parovic-etal-2022-bad, parovic-etal-2023-cross, rathore-etal-2023-zgul, moomin-2024-modular, dementieva-etal-2025-cross}. This is in part because they are robust to hyperparameter choices and training data amounts, in contrast to full model fine-tuning \citep{pfeiffer2024modular}.
Some PEFT methods can successfully use monolingual data to improve MT performance; for example, \citet{ustun-etal-2021-multilingual} train denoising adapters for both source and target languages, inserting the adapters between the encoder and decoder layers of an MT model, respectively.
By freezing all model weights except for the cross-attention of the decoder, they then carry out cross-attention fine-tuning (CA-FT) with only a limited amount of parallel data. 
\citet{chronopoulou-etal-2023-adaptersoup} also address data scarcity in adaptation to new domains following \citet{wortsman-etal-2022-model} by ``souping'' (\textit{i.e.,} weight-space averaging) pretrained domain adapters, as in Equation \ref{eq:soup} (where $\theta_i$ represent the weights of each domain adapter).
We combine the principles of these prior works in our paper.

\begin{equation}\label{eq:soup}
    \theta_{soup} = \frac{1}{l} \sum_{i=1}^{l} \theta_i
\end{equation}

Prior work has also catalyzed transfer by training language family adapters to leverage phylogenetic relations between languages \citep{faisal-anastasopoulos-2022-phylogeny,chronopoulou-etal-2023-language}.
Other factors of selecting transfer languages have been investigated in the past besides phylogeny.
These include similarity in terms of explicit grammatical features like word order \citep{pires-etal-2019-multilingual}, similarity in terms of language representation space \citep{ustun-etal-2022-hyper}, as well as lexical overlap \citep{pfeiffer-etal-2021-unks,de-vries-etal-2022-make}.
Other metrics have been proposed to select candidate transfer languages, including Subword Evenness (SuE; \citealp{pelloni-etal-2022-subword}) which quantifies how uneven subword lengths are in a multilingual tokenizer: uneven implies successful transfer.

In this work we focus on MT for Creole languages, a set of low-resource languages with hundreds of millions of speakers who presently lack reliable MT systems, despite demonstrable need for them \cite{neto2020latin,robinson-etal-2024-kreyol}. 
Creole language technologies have not been thoroughly researched or developed; yet several works that do exist inspired our decision to explore them (cf.~our literature review in Appendix \ref{app:lit}).
Although most Creoles are genealogically related to high-resource languages (\textit{e.g.}, English, French, Spanish), previous works have had mixed success leveraging transfer learning to boost performance for Creoles \cite{lent-etal-2022-ancestor,lent-etal-2024-creoleval,robinson-etal-2022-data,robinson2023african}.
No prior work has conclusively determined which languages to select for successful Creole transfer, explored PEFT in this application, nor compared linguistic and non-linguistic language relations in Creole transfer language selection. 
These aspects combined make Creoles a viable testbed for comparing transfer criteria.

We apply a new adapter method inspired by \citeposs{ustun-etal-2021-multilingual} approach to MT for three Creole languages. 
(See details in \S \ref{sec:meth_exp}.) 
We experimented with this approach to compare a large variety of dimensions of linguistic relatedness and found that, contrary to our initial hypotheses, gains are virtually indistinguishable regardless of which transfer criterion we apply. 
Still more surprisingly, incorporating unrelated languages or noise to the Creole adapters yields comparable improvements. 
We conjecture that even untrained adaptors may assist cross-attention fine-tuning by acting as regularizers. 
We couple this hypothesis with analysis presented in \S \ref{sec:res_ana}.

We contribute both \circled{1} a novel regularization method combining adapter souping with cross-attention fine-tuning that surpasses baselines by 8 BLEU in some settings, and \circled{2} evidence that some of the benefit of adapters lies in regularization, independent of a wide range of linguistic information.

\section{Methodology and Experiments} 
\label{sec:meth_exp}

We follow \citet{ustun-etal-2021-multilingual} using their two-step methodology.
First, we train monolingual denoising adapters separately for each language.
At inference time, we add the relevant adapters to the encoder and decoder corresponding to the source and target languages, and fine-tune the decoder cross-attention on small amounts of parallel bitexts.
Furthermore, we induce transfer by weight-averaging ("souping") Creole adapters with candidate transfer language adapters inspired by previous efforts \citep{faisal-anastasopoulos-2024-data,wortsman-etal-2022-model,chronopoulou-etal-2023-adaptersoup}.
The novelty lies in combining source and target language adapters, cross-attention fine-tuning and souping in a single method for improving low-resource MT.
(See more details about our methodology in Section \ref{subsec:exp}).

We focus our efforts on three Creole languages: Haitian, Papiamento, and Sango, spoken in total by roughly 18 million speakers in Haiti, the ABC Islands, and the Central African Republic, respectively.\footnote{According to \href{https://www.ethnologue.com}{Ethnologue}.}
This selection is primarily due to the intersection of data availability and support by NLLB-200, the MT model we apply in experiments \citep{nllb2022}.
Despite their unique contexts, these languages share commonalities in socio-historical development: they evolved from linguistic contact between European and African languages resulting from Western colonialism \cite{kouwenberg2009handbook}.
As such, they can trace parts of their lexicon and grammar back to both European and African languages. 

\subsection{Selecting Languages for Transfer}

We examine five principled approaches to select transfer languages, which range from linguistically motivated to purely data-driven.
First, we leverage language phylogeny, selecting likely Indo-European (\textbf{IE}; \citealp{rickford1997language, lefebvre2011substrate}) and Niger-Congo (\textbf{NC}; \citealp{glottolog,mufwene2006pidgins,hall1958creole,faine1937philologie}) \textbf{relatives} for Haitian, Papiamento, and Sango.
We also experiment with transfer between these  \textbf{Creole languages}, as previous works have argued that Creole languages have linguistic commonalities with each other \cite{daval-markussen-bakker-2012-explorations}.  
We next select languages by similarity of documented typological features, calculated as \textbf{\texttt{lang2vec}} distances \citep{littell-etal-2017-uriel}. 
For data-driven language selection, we determine language relatedness from similarity of NLLB-200 language embeddings (\textbf{NLLB representations}; \citealp{nllb2022}), 
as well as Subword Evenness (\textbf{SuE}; \citealp{pelloni-etal-2022-subword}). 
For details of our language selection processes, see Appendix \ref{app:selection}. 

\subsection{Control experiments}

As an experimental control, we identify three language groups with no linguistic relation to the Creole languages in question: \textbf{Uralic} languages, \textbf{Dravidian} languages, and \textbf{CJK} (\textit{i.e.}, Chinese, Japanese, and Korean). 
None of these groups share phylogeny or significant amounts of typological features with our Creole languages. 
The latter two also differ in terms of script. 
And the final group is distinct in that its languages also bear no phylogenetic relation to each other. 
As a final experimental control, we use a randomly initialized, untrained adapter (\textbf{\texttt{init}}). 
We use this untrained adapter in two ways: as a replacement for the Creole language adapter, and souped with the Creole language adapter. 
As \citet{chronopoulou-etal-2023-adaptersoup} suggest for souping, we use the same random initialization for all untrained and trained adapters.

\subsection{Experimental Details}\label{subsec:exp}

\begin{table}[]
    \centering
    \resizebox{\columnwidth}{!}{
    \begin{tabular}{lll}
    \toprule
    Dataset & Domain & Source \\
    \midrule
    MADLAD & Web scrape & \citealp{kudugunta2023madlad400} \\
    NLLB-OPUS & Web scrape & \citealp{nllb2022} \\
    FLORES-200 & Wikipedia & \citealp{nllb2022} \\
    \bottomrule
    \end{tabular}
    }
    \caption{
    Datasets and domains used. We use MADLAD to train adapters, NLLB-OPUS to fine-tune cross-attention, and FLORES-200 to evaluate.
    }
    \label{tab:data}
\end{table}

Our experiments follow the same general steps.
\circled{1} We train all language adapters individually on denoising using monolingual data from MADLAD-400 \citep{kudugunta2023madlad400} that we preprocess and clean further.\footnote{See Appendix \ref{app:datapreproc} for preprocessing details.}
\circled{2} We add the source language adapter $s$ to the encoder and the target language adapter $t$ to the decoder and freeze all adapter and model parameters.
In adapter averaging experiments the \texttt{init} or the helper language adapters are souped with the $s$ or $t$ language adapters in this step as well.
\circled{3} We unfreeze the cross-attention of the decoder and use 10k parallel data from NLLB-OPUS \citep{nllb2022} to do cross-attention fine-tuning (CA-FT).
\circled{4} We evaluate MT from $s \rightarrow t$ on Flores-200 \citep{nllb2022} using mean BLEU and chrF scores. (See Table \ref{tab:data} for data composition.)

All experiments are conducted using a 600M-parameter distillation of the original 54B-parameter NLLB-200 model \citep{nllb2022} which has 12 encoder and 12 decoder layers, 16 attention heads, 1,024 dimensions, and a model vocabulary of over 250,000 sentence-piece tokens shared across all languages.
We adopt the adapter training and the hyperparameters from \citet{ustun-etal-2021-multilingual}, training bottleneck adapters on 10,000 training samples for a maximum of 100,000 training steps with early stopping, a batch size of 8, 4,000 warmup steps, and the Adam optimizer with a maximum learning rate of $2e-4$ with a gradient accumulation of 8 steps.
All adapters initialized from the same random seed as \citet{chronopoulou-etal-2023-adaptersoup} show that this is necessary for successful souping.
We train CA-FT for 1 epoch with early stopping.
We soup adapters with the same weight except when averaging with the \texttt{init} adapter where the weight of the Creole language to \texttt{init} adapter is $1:3$, to closer approximate the other setups where the Creole language adapter is souped with three other adapters.

\section{Results and Analysis}
\label{sec:res_ana}

Here we present only Creole-to-English BLEU scores. 
We conducted preliminary experiments using only 800 aligned fine-tuning segments on these and the inverse translation directions, and saw the same trends shown in our results here. 
We decided to prioritize into-English MT for our main experiments after seeing small adapter gains across the board for into-Creole. 
(See Appendix~\ref{app:other} for our full results suite.) 
Table \ref{tab:results} shows BLEU improvements of our approaches over both baselines: the base MT model with CA-FT, and the $s$ and $t$ language adapters without souping \citep{ustun-etal-2021-multilingual}. 
The improvements, however, are not particularly differentiated between various transfer and control scenarios, and notably, untrained adapters match or exceed performance of trained ones.

\begin{table}[]
    \centering
    \resizebox{\columnwidth}{!}{
    \begin{tabular}{lrrr}
    \toprule
    Experiment & hat $\rightarrow$ eng & pap $\rightarrow$ eng & sag $\rightarrow$ eng
    \\
    \midrule
    Base Model (CA-FT) & 33.37 & 38.97 & 10.89 \\
     $s$ and $t$ Adapters & 32.33 & 40.04 & 11.40 \\
    \includegraphics[scale=0.02]{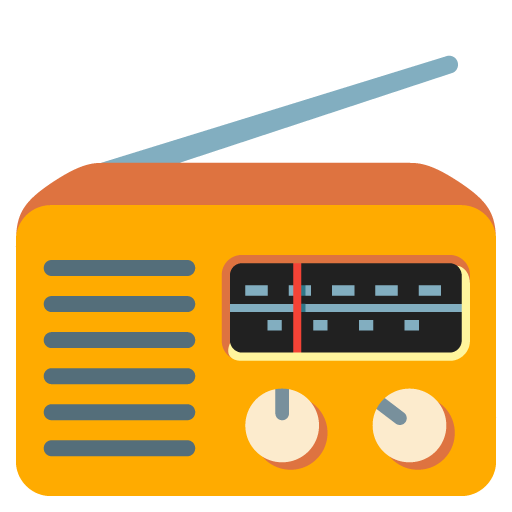} Untrained $s$ Adapter & 37.07 & 45.01 & 14.91 \\
    \midrule
    \midrule
    \includegraphics[scale=0.02]{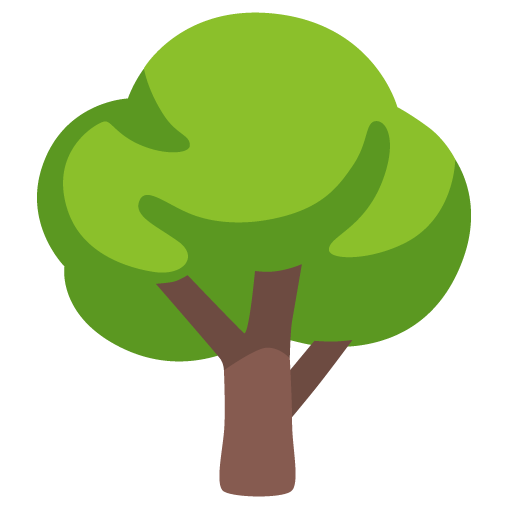} IE Transfer & 36.44 & 46.35 & 12.46 \\
    \includegraphics[scale=0.02]{emoji/tree.png} NC Transfer & 36.06 & 46.69 & 12.29 \\
    \includegraphics[scale=0.02]{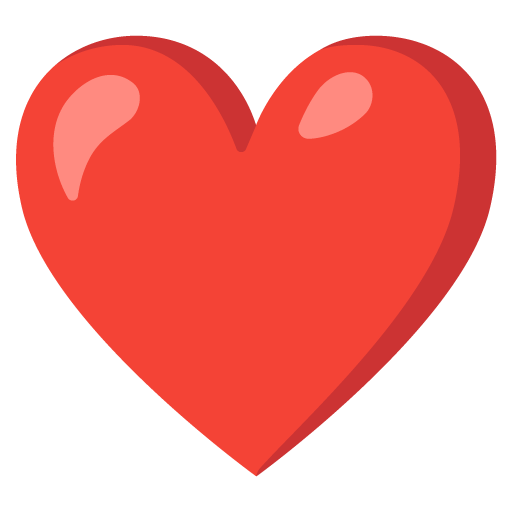} Creole Transfer & 35.25 & 46.23 & 12.76 \\
    \includegraphics[scale=0.02]{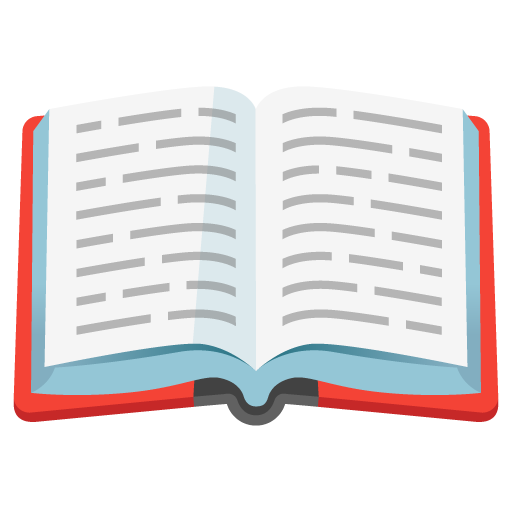} lang2vec & 36.54 & 47.04 & 13.07  \\
    \includegraphics[scale=0.02]{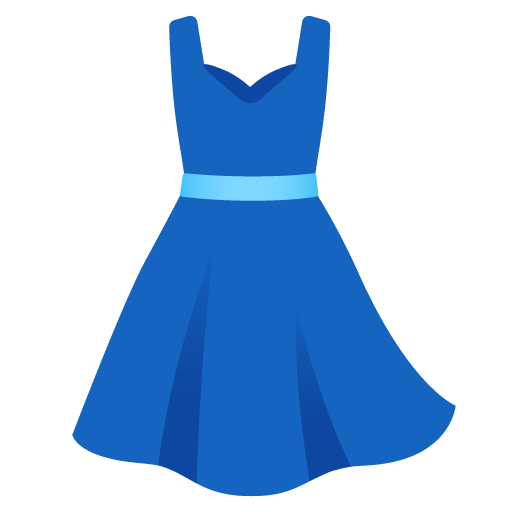} NLLB Vec & 35.80 & 46.91 & 12.80  \\
    \includegraphics[scale=0.02]{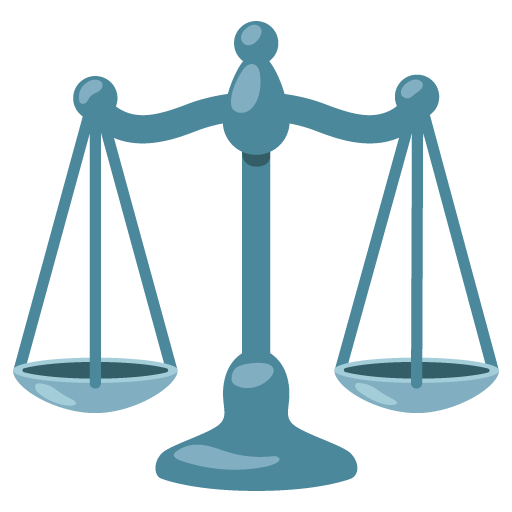} Subword Evenness & 36.36 & 47.03 & 13.12 \\
    \midrule
    Mean Improvement & \cellcolor{green!30!yellow} $+2.70$ & \cellcolor{green} $+7.74$ & \cellcolor{green!15!yellow} $+1.86$ \\
    \midrule
    \includegraphics[scale=0.02]{emoji/radio.png} Untrained Souping & 37.42 & 46.34 & 13.41 \\
    \includegraphics[scale=0.02]{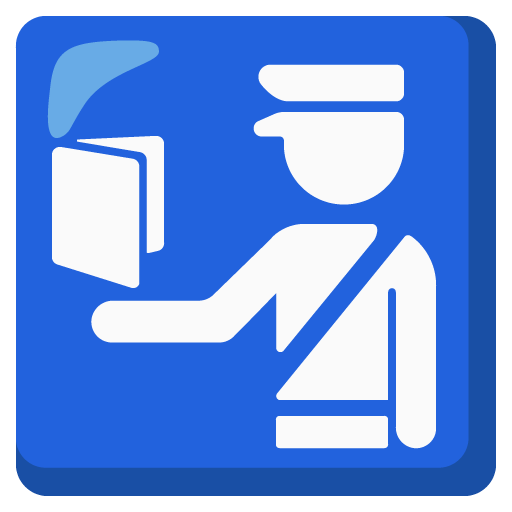} Uralic & 37.06 & 47.00 & 13.58 \\
    \includegraphics[scale=0.02]{emoji/control.png} CJK & 36.41 & 47.17 & 13.33 \\
    \includegraphics[scale=0.02]{emoji/control.png} Dravidian & 36.55 & 47.27 & 13.27 \\
    \midrule
    Mean Improvement & \cellcolor{green!30!yellow} $+3.49$ & \cellcolor{green} $+7.98$ & \cellcolor{green!30!yellow} $+2.51$ \\
    \bottomrule
    \end{tabular}
    }
    \caption{
    Mean BLEU scores for Creole to English MT experiments over a diverse set of conditions for selecting transfer languages, as motivated by phylogeny (\includegraphics[scale=0.02]{emoji/tree.png}), language group (\includegraphics[scale=0.02]{emoji/heart.png}), typological databases (\includegraphics[scale=0.02]{emoji/book.png}), model representations (\includegraphics[scale=0.02]{emoji/dress.png}), and tokenization (\includegraphics[scale=0.02]{emoji/scale.png}).
    Experimental results are also compared against control experiments with unrelated languages (\includegraphics[scale=0.02]{emoji/control.png}) and untrained adapters (\includegraphics[scale=0.02]{emoji/radio.png}).
    We highlight mean performance improvements over the Base Model (CA-FT) for both transfer and controls.
    }
    \label{tab:results}
\end{table}

\paragraph{Regularization, Not Transfer?}

Two facets of our results indicate that cross-lingual transfer does not explain the success of our methodology.
The first is that adapter souping improves performance over the baselines, \textit{regardless} of the criteria used to select transfer languages. The unrelated control languages (\textit{i.e.}, Dravidian, Uralic, and CJK) achieve performance gains on par with languages selected through principled means (\textit{e.g.}, genealogy and typology), without marked differences in BLEU or chrF scores. 
This holds for both the source and target language adapters in the encoder and decoder, respectively.
The second is the commensurate success of the untrained adapters in averaging. Souping adapters with `meaningful' language adapters is no better than souping with random noise, indicating that cross-lingual transfer (or any information transfer) is likely not responsible for performance gains. 
These findings seem more consistent with previous works on regularization, which find that adding noise can improve a model's ability to generalize and avoid overfitting \citep{hochreiter-schmidhuber-1994-simplifying, poole-etal-2014-analyzing, moradi_survey_2020} (which may be pertinent given both our limited train sets and train/test domain mismatch). 

We perform multiple analyses to test our hypothesis that adapters are regularizing. 
First, we compared the use of untrained adapters to the use of adapters of known closely related languages. 
Because the close relatives of Creole languages are disputed, we turned to another relatively low resource language, Catalan, which has three close relatives also supported by NLLB-200. 
Spanish, Portuguese, and Occitan are closely related to Catalan by effectively every reasonable linguistic criterion. 
Table \ref{tab:catalan} shows different trends for Catalan (a language on which NLLB's baseline performance is notably high). 
In this scenario, none of the adapter approaches improve performance over the baseline, once again indicating that the monolingual denoising adapter training on MADLAD data is not positively contributing. 
And once again, even with these highly attested close linguistic relationships, we see the random adapter comes nearest to recovering the baseline score in both data settings. 
This set of results underscores the possibility that CA-FT without randomized regularization may lead to overfitting (hence the performance drops). 
We also experimented souping only one or two related languages for both Catalan$\rightarrow$English and Papiamento$\rightarrow$English but saw no improvements.

\begin{table}[]
    \centering
    \resizebox{\columnwidth}{!}{
    \begin{tabular}{lrr}
    \toprule
    FT examples: & 800 & 10k 
    \\
    \midrule
    Base Model (CA-FT) & 45.45 & 45.53 \\
    \textit{s} and \textit{t} Adapters & 38.58 & 41.92 \\
    \midrule
    \includegraphics[scale=0.02]{emoji/tree.png} \texttt{spa} + \texttt{por} + \texttt{oci} Souping & 41.87 & 43.74 \\
    \includegraphics[scale=0.02]{emoji/radio.png} Untrained Souping & 43.97 & 44.75 \\
    \bottomrule
    \end{tabular}
    }
    \caption{
    BLEU scores for Catalan to English MT.
    }
    \label{tab:catalan}
\end{table}

\begin{figure}
    \centering
    \includegraphics[width=0.9\columnwidth]{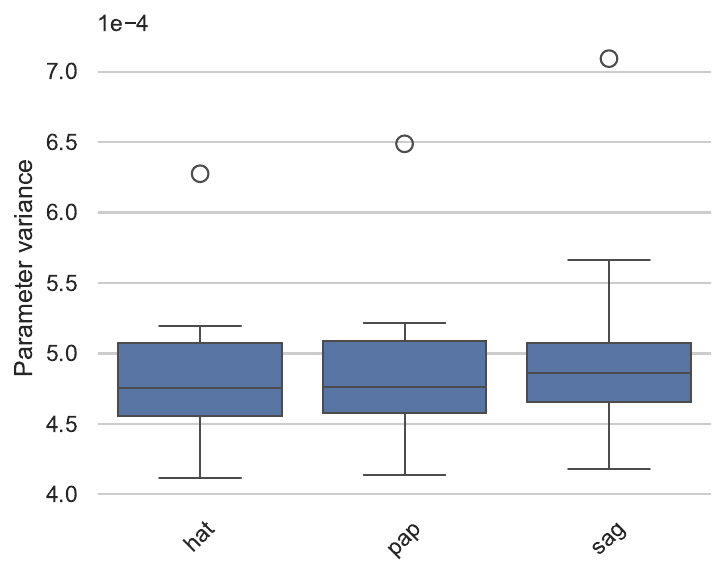}
    \caption{
    Parameter variance between the pretrained Creole adapters (represented by the outliers) vs the souped adapters of various experimental conditions (shown in the boxplots).
    }
    \label{fig:parameter_var}
\end{figure}

We observed training loss, gradient norms, and dev loss for CA-FT of the experiments presented in this section. 
We compare untrained adapter souping (hypothesized regularization) with simple \textit{s} and \textit{t} adapters (unregularized), finding that the unregularized case has moderately higher gradient norms and dev loss, consistent with our hypothesis. 
The unregularized case also sees higher train loss, which is inconclusive regarding our hypothesis since an overfit model may be stuck in a local minimum or be overfit on a distribution varying from the CA-FT data. 
(See Appendix \ref{app:loss}.)
We also show that parameter variance decreases substantially when souping adapters compared to the pretrained Creole adapters (Figure \ref{fig:parameter_var}), further supporting our regularization explanation.

\paragraph{Qualitative Evaluation}

Beyond computational analyses, we further assess the quality of translations across our experimental conditions.
To start, a subset of the authors create a simple annotation guideline, inspired by MQM \cite{burchardt-2013-multidimensional}, to help identify errors regarding vocabulary, grammar, and other translation issues (\textit{e.g.}, addition or omission). 
Then, one author, a native Haitian speaker, carries out a linguistically-grounded manual evaluation of Haitian $\rightarrow$ English translation across the baseline of $s$ and $t$ Adapters, the IE Transfer, and the Untrained Souping.
The evaluation involves 33 samples per condition presented to the annotator in three shuffled sets (\textit{i.e.}, 99 sentences in total) to avoid biasing him.
The primarily focus falls on grammar and translation issues
(For a condensed list of errors, see Table \ref{tab:mqm_errors} in Appendix \ref{app:mqm_errors}).
Although a small sample size, we observe that adapter souping results in less grammatical and lexical errors when applying the pretrained Haitian and English adapters with CA-FT.
While both IE Transfer and Untrained Souping are comparable in reducing lexical errors, the latter does better in avoiding grammatical errors.
This may indicate a comparative success of regularization when adding random noise that is also reflected in the $+0.98$ BLEU score Untrained Souping has over IE Transfer for Haitian $\rightarrow$ English (see Table \ref{tab:results}).
We believe that even a small manual translation evaluation demonstrates the value of involving native speakers in MT.

\section{Conclusion}

Inspired by prior success of monolingual adapters and CA-FT for low-resource MT, we corroborate prior findings that adapters can help by consistently improving BLEU scores for a set of Creole languages. 
However, we differ from prior research by finding no evidence of linguistic or informational transfer from adapters, even across a wide array of linguistic relationships. 
We hypothesize that most adapter aid in this setting is from regularization and find some analytical support for this. 

\section*{Limitations}

\paragraph{Scalability} 

This work can only be applied to a small subset of all Creoles that are present in larger benchmarks \cite{lent-etal-2024-creoleval, robinson-etal-2024-kreyol} due to the fact that they are low-resource languages -- in certain cases, extremely so --, so there is an issue of data scarcity and a limited support by MT models.
This data scarcity extends to also the African relatives of the Creoles we target in this paper.
Our work is further limited by the limited coverage by typological resources.
Grambank \citep{grambank_release} supports only Haitian, not Papiamento and Sango, and while the lang2vec covers more languages, the coverage is limited and uneven.
Despite these constraints, however, we believe our selection of Creoles (\textit{i.e.}, Haitian, Papiamento, and Sango) are diverse enough to make high-level hypotheses about transfer learning for Creoles; these languages have different relatives (\textit{e.g.}, inheriting large vocabulary from French, Portuguese, and Ngbandi, respectively) and are spoken across different regions (\textit{i.e.}, the Caribbean and Africa). Moreover, our
 experimentation over these languages is \textbf{extensive}, with careful investigation of 6 transfer scenarios, 1 \texttt{init} baseline, and 3 principled controls. So while this work cannot scale \textit{widely} across Creoles, we investigate transfer scenarios for each Creole \textit{deeply}.

\paragraph{Domain of Data}
Similar to many other low-resource languages, much of the available data for Creoles comes from the religious domain (\textit{e.g.}, Bible translations), which has a characteristic style that is typically not representative of language use in other contexts \cite{mielke-etal-2019-kind}. 
Moreover, religious domain content may not necessarily be culturally appropriate for some communities \cite{hershcovich-etal-2022-challenges, mager-etal-2023-ethical}.
In order to be truly useful for end users, data sourced from a wide variety of domains will ultimately be a necessary component for the development of robust MT systems for Creoles. 
Until more such data becomes available, we believe that work at the intersection of transfer learning and domain adaptation could be greatly beneficial for Creole NLP. 

\paragraph{Difficulty in Verifying Neural Phenomena} 
While we find some analytical evidence that our adapters are boosting performance by regularizing fine-tuning, it is difficult to verify this with certainty. 
Regularization is a mathematical phenomenon that can be demonstrated rigorously, but such involved analysis is outside the scope of this current work; hence we leave these important analytical details to future work.

\section*{Ethics Statement}

As Creole languages are often marginalized, we recognize the necessity of heightened ethical scrutiny for research involving these languages. We thus engage directly with considerations from \citet{lent-etal-2022-creole}, \citet{mager-etal-2023-ethical}, and \citet{bird-2020-decolonising}, to ensure that our work is in accordance with the needs and desires of speakers. To this end, we note that speakers of many different Creole languages actively \textit{want} and/or \textit{need} improved MT technology for their language. 
In the context of Haitian Creole, for example, MT technology is highly desirable for facilitating interactions with non-Haitian speakers (\textit{e.g.}, the neighboring, Spanish-speaking Dominican Republic, and foreign workers at NGOs operating in Haiti). Additionally, MT has the potential to greatly improve access to information for Haitians. \citet{degraff2022language} argues that the lack of access to education in Creole is a critical barrier to social and economic mobility in Haiti. Accordingly, reliable MT for Haitian Creole could make available high-quality educational resources from platforms such as edX, Coursera, MITx, and Udemy, for example. Thus, if MT systems could offer accurate translations into Haitian Creole, they would become transformative tools for advancing the country's development, breaking down educational barriers, and fostering greater social equity. We note similar potential for MT for speakers of Sango in Central African Republic, Chad, and Democratic Republic of the Congo and \cite{afdb2024}. And while the socioeconomic context of Aruba, Bonaire, and Curacao is quite different from that of aforementioned Haiti or Central African Republic, speakers' desire for language technology for Papiamento is well documented through projects like \href{https://www.papiamentu.ai/}{papiamentu.ai}.

\section*{Acknowledgments}
MF, JB, and HL were supported by 
the Carlsberg Foundation, under the \textit{Semper Ardens: Accelerate} program (project nr. CF210454); the same project also supported the annotation work performed by DJB. 
EL was funded by the Google Award for Inclusion Research program (awarded to JB and HL for the “CREOLE: Creating Resources for Disadvantaged Language Communities” project).
We extend our thanks to Mike Zhang, Esther Ploeger, and the other members of AAU-NLP, for their helpful feedback on our manuscript.

\bibliography{custom,anthology}

\appendix

\section{Literature Review}\label{app:lit}

NLP research for Creole languages is still in relatively early stages. 
Some large-scale multilingual MT systems have limited Creole language support: NLLB \cite{nllb2022} supports five Creole languages, and Google Translate\footnote{\url{https://translate.google.com/} as of 8 Feb., 2025} and Lego-MT \cite{yuan-etal-2023-lego} each support eight.  
Other works have focused specifically on building MT for Creole languages Haitian \cite{robinson-etal-2022-data,callison-burch-etal-2011-findings,frederking-etal-1997-diplomat}, Singlish \citep{wang-etal-2017-universal,liu-etal-2022-singlish}, Nigerian Pidgin \citep{adelani-etal-2022-thousand,ogueji2019pidginunmt}, Mauritian Creole \citep{dabre-sukhoo-2022-kreolmorisienmt}, Sranan Tongo \citep{zwennicker2022towards}, Guyanese Creole \cite{clarke-etal-2024-guylingo}, and Jamaican Patois \cite{robinson-etal-2022-data}. 
Some others have focused on building Creole language-specific NLP tools outside of MT for Nigerian Pidgin \citep{muhammad-etal-2023-semeval,muhammad-etal-2022-naijasenti,adelani-etal-2021-masakhaner,oyewusi2020semantic,caron-etal-2019-surface}, Jamaican Patois \citep{armstrong-etal-2022-jampatoisnli}, and Martinican Creole \citep{mompelat-etal-2022-parse}.  

More recently, researchers have developed large scale multi-Creole-language NLP applications. 
\citet{lent-etal-2024-creoleval} developed CreoleVal, a multitask benchmark for Creoles including an MT dataset and model for 28 Creole languages. 
\citet{robinson-etal-2024-kreyol} expanded on \citeposs{lent-etal-2024-creoleval} work, building genre-diverse MT datasets and models for 41 Creole languages of the Caribbean, Latin America, and Colonial Africa. 
Perhaps most pertinent to our current work are past studies into cross-lingual transfer to benefit Creole NLP. 
In tightly controlled experiments, \citet{robinson2023african} found indications that Haitian and Jamaican Patois' Niger-Congo relatives may be more helpful as transfer languages than their Indo-European relatives, for MT applications. 
\citet{lent-etal-2022-ancestor}, however, did not identify clear advantages from a language's relatedness to a Creole language in terms of its usefulness in transfer learning. 
Given the sparse research and even sparser positive results in the area, optimal transfer learning for Creole NLP remains an open question.  

\section{Data Preprocessing}\label{app:datapreproc}

The MadLad corpus is organized in series of internet-scraped documents, which the MadLad team cross-checks for noise \cite{kudugunta2023madlad400}. For training the monolingual adapters, we use the \textit{MadLad-Clean} subsplit and we perform further preprocessing, since we noted after manual inspection that noise can still be present in the documents. Firstly, the MadLad documents are split into sentences using the SentenceSplitter \footnote{https://github.com/mediacloud/sentence-splitter} library.

For each sentence in the document, we define the following rules, if any of the rules is breached, then the given sentence is not included in the final monolingual training corpus.

\begin{enumerate}[noitemsep,topsep=1pt]
    \item Sentence ends with a terminal punctuation mark ('.'; '!', ' " ', '?').
    \item Ratio of numerical characters with respect to all other characters is less than \textbf{0.25}.
    \item Sentence starting with a bullet-point.
    \item Sentence containing a \textit{lorem ipsum} substring.
    \item Ratio of stop-words with respect to all other tokens is less than \textbf{0.6} using whitespace tokenization.
    \item Ratio of token count that does not contain alpha characters with respect to total token count is less than \textbf{0.25}, using whitespace tokenization.
    \item Sentence length is less than \textbf{14} and more than \textbf{300} characters.
    \item Ratio of capitalized characters is less than \textbf{0.2}.
    \item Ratio of alpha characters with respect to all other characters is more than \textbf{0.6}.
    \item Language identification does not match the expected language. For this we employ FastText model \footnote{https://huggingface.co/facebook/fasttext-language-identification}.
\end{enumerate}

This preprocessing removed approximately ~20-23\% of sentences, depending on the language. The same preprocessing steps were also applied to NLLB parallel data used for cross-attention fine-tuning, where preprocessing removed approximately 30-50\% of the sentences.

\section{Selecting Languages for Transfer}\label{app:selection}

\begin{table}[]
    \centering
    \begin{tabular}{ll}
    \toprule
    Condition & Language \\
    \midrule
    \midrule
    \multicolumn{2}{c}{Haitian} \\
    \midrule
    \midrule
    Creole Language & hat \\
    IE Relative & fr, es, en \\
    NC Relative & ig, ee, fon \\
    \texttt{lang2vec} & grc, oc, fo \\
    NLLB & pap, epo, ltz \\
    \midrule
    \midrule
    \multicolumn{2}{c}{Papiamento} \\
    \midrule
    \midrule
    Creole Language & pap \\
    IE Relative & nl, pt, es \\
    NC Relative & kg, ak, fon \\
    \texttt{lang2vec} & suz, sr, ug \\
    NLLB & hat, war, tgl \\
    \midrule
    \midrule
    \multicolumn{2}{c}{Sango} \\
    \midrule
    \midrule
    Creole Language & sag \\
    IE Relative & fr \\
    NC Relative & kbp \\
    \texttt{lang2vec} & ig, yo, ca \\
    NLLB & kg, tsn, lin \\
    \midrule
    \midrule
    Subword Evenness & kat, rus, zsm \\
    \midrule
    \midrule
    \multicolumn{2}{c}{\textbf{Controls}} \\
    \midrule
    \midrule
    Uralic & fin, est, hun \\
    CJK & zho, jap, kor \\
    Dravidian & te, ta, ka \\
    \midrule
    \midrule
    \multicolumn{2}{c}{\textbf{Other Settings}} \\
    \midrule
    \midrule
    \multicolumn{2}{c}{Hungarian} \\
    \midrule
    \texttt{lang2vec} & fin, tur, als \\
    NLLB & fin, bul, hrv \\
    \midrule
    \multicolumn{2}{c}{Yoruba} \\
    \midrule
    \texttt{lang2vec} & tha, ibo, fon \\
    NLLB & ibo, hau, xho \\
    \midrule
    \midrule
    \bottomrule
    \end{tabular}
    \caption{All languages used for the various experimental settings. 
    }
    \label{tab:all_languages}
\end{table}

Table~\ref{tab:all_languages} displays the different transfer languages we selected in our experiments. 
Below we detail our processes of selecting them. 

\subsection{Phylogeny-based Transfer}

\citet{faisal-anastasopoulos-2022-phylogeny} followed principles of language phylogenetics to inform their transfer learning experiments, training adapters on combined Indo-European data, and for subfamilies of Indo-European such as Germanic, Romance, Indo-Aryan, etc. 
To translate for French, for example, they stacked a French adapter on top of a Romance and then an Indo-European adapter at inference time. 
Creole phylogeny is heavily disputed, however \cite{kouwenberg2009handbook}. Linguists have debated whether Atlantic Creole languages such as Haitian and Papiamento are best understood as Niger-Congo (NC) languages with extensive Indo-European (IE) lexical borrowings \citep{lefebvre2011substrate,rickford1997language}, or as IE languages that developed with strong NC influence \citep{faine1937philologie, hall1958creole, glottolog, mufwene2006pidgins}. 
Considering this lack of consensus, we select two groups of related languages for each Creole language: IE relatives and NC relatives. 
In this work, we made a selection based on APICS \cite{apics} and linguistic references. %

\subsection{Typological Feature-based Transfer}

Information about the typological features of languages is typically documented in databases such as APICS \cite{apics}, WALS \cite{wals}, and Grambank \cite{grambank_release}.
The \texttt{lang2vec} library \citep{littell-etal-2017-uriel} assembles typological feature vectors from different sources, enabling measurement of holistic typological distance between languages.
The drawbacks of \texttt{lang2vec} include the lack of equal coverage of languages in terms of the number of features, and the coarseness of the linguistic properties \citep{ponti-etal-2019-modeling}.
To supplant the lack of coverage, we use \texttt{data2lang2vec} which uses textual data to inform feature-prediction of the missing features \citep{amirzadeh-etal-2025-data2lang2vec}.

\subsection{Data-driven Language Selection Methods}

\paragraph{NLLB Language Embeddings} Typological similarity between languages can also be derived in a data-driven manner, \textit{e.g.}, from language embeddings.
To derive the similarity between languages this way, we utilize the language embeddings of the distilled NLLB-200 model which implicitly contain much information with regards to the various languages especially with regards to morphosyntax and vocabulary. 

\paragraph{Subword Evenness} Introduced by \citet{pelloni-etal-2022-subword}, Subword Evenness (SuE) is a metric quantifying the differences in the length of subword units for each individual language. It presents another data-driven approach to transfer language selection. 
In this metric, lower SuE scores have been demonstrated as a good predictor of transfer success to various low-resource languages.
Thus, we select the 5 languages in the MADLAD dataset with the lowest SuE angles as calculated when using a BPE tokenizer, see Table \ref{tab:sue}. 

\begin{table}[h]
    \centering
    \resizebox{\columnwidth}{!}{
    \begin{tabular}{lllr}
    \toprule
    Language Code & Language Code & Script & SuE Angle \\
    \midrule
    $\prec$azb & South Azerbaijani & Arabic & 66.72 \\
    kat & Georgian & Georgian & 72.57 \\
    zsm & Malay & Latin & 79.80 \\
    $\prec$kas & Kashmiri & Devanagari & 80.79 \\
    rus & Russian & Cyrillic & 81.21 \\
    \bottomrule
    \end{tabular}
    }
    \caption{MADLAD Languages with the lowest SuE angles calculated using a BPE tokenizer.
    We note that MADLAD only has Norther Azerbaijani (Latin script), and not South Azerbaijani (Arabic script). 
    Similarly, Kashmiri is only in MADLAD with the Arabic script, not with Devanagari.
    }
    \label{tab:sue}
\end{table}

\subsection{Control Experiments with Unrelated Languages}
\label{sec:ctrl}

To determine whether improvements over Creoles are truly the result of successful cross-lingual transfer, 
we compare the previously described methods against an equivalent adapter approach using entirely unrelated languages. 
We use three different sets of such unrelated languages.
The first set is made up of the \textbf{Uralic languages} Finnish, Estonian, and Hungarian, which are written with the Latin script but have no phylogenetic relation to Haitian, Papiamento, and Sango.
The second set consists of Chinese, Japanese, and Korean -- so-called \textbf{CJK languages} --, which belong to three distinct language families and are written with three distinct scripts having little to no overlap with Atlantic and African Creole languages.
The third group consists of the \textbf{Dravidian languages} Tamil, Telugu, and Kannada, which are closely related to each other but also written with three distinct scripts, and again are unrelated to the Creole languages in terms of phylogeny.

\section{Full Results}
\label{app:other}

See Table \ref{tab:tgt_avg_flores_old} for our results translating between English and Creole languages using only 800 fine-tuning parallel segments.

\begin{table*}[ht]
    \centering
    \resizebox{\textwidth}{!}{
    \begin{tabular}{lrrrrrr}
    \toprule
    Experiments & en $\rightarrow$ hat & en $\rightarrow$ pap & en $\rightarrow$ sag & hat $\rightarrow$ en & pap $\rightarrow$ en & sag $\rightarrow$ en \\
    \midrule
     \multicolumn{7}{c}{Baselines} \\
     \midrule
    dNLLB-200 w/ CA-FT & 23.59 & 23.56 & 11.16 & 33.36 & 39.05 & 10.62 \\
    \midrule
    \multicolumn{7}{c}{Language Adapters} \\
    \midrule
    Source + Target LA & 23.55 & 26.82 & 11.90 & 31.18 & 38.23 & 12.19 \\
    \midrule
    \multicolumn{7}{c}{Phylogeny-Inspired Adaptation} \\
    \midrule
    Target + IE Relatives & 24.01 & 29.35 & 11.94 & 35.24  & 45.48  & 13.51  \\
    Target + NC Relatives & 23.98 & 29.71 & 11.93 & 34.96  & 46.13  & 13.03 \\
    \midrule
    \multicolumn{7}{c}{Typological Adaptation} \\
    \midrule
    \texttt{lang2vec} & \textbf{24.28}  & \textbf{30.21}  & 11.71 & \textbf{35.63}  & \textbf{46.95}  & 13.79  \\
    Target + Data-Driven (NLLB) & 24.09 & 30.17 & \textbf{12.01} & 34.75  & 46.09  & 13.48 \\
    \midrule
    \multicolumn{7}{c}{Miscellaneous} \\
    \midrule
    Creole Transfer & 23.99 & 29.87 & 11.71 & 34.24 & 44.75 & 13.48 \\
    Subword Evenness & 24.23 & 29.73 & 11.80 & 35.20  & 46.20  & \textbf{14.11} \\
    \midrule\midrule
    Largest Improvement from dNLLB & \cellcolor{lightyellow} $+0.69$ & \cellcolor{green} $+6.65$ & \cellcolor{lightyellow} $+0.85$ & \cellcolor{yellow} $+2.27$ & \cellcolor{green} $+7.90$ & \cellcolor{green} $+3.49$ \\
    Largest Improvement from Üstün & \cellcolor{lightyellow} $+0.73$ & \cellcolor{green} $+3.39$ & \cellcolor{lightyellow} $+0.11$ & \cellcolor{green} $+4.45$ & \cellcolor{green} $+8.63$ & \cellcolor{yellow} $+1.92$ \\
    \midrule
    \multicolumn{7}{c}{Control} \\
    \midrule
    Uralic & 24.31 & 30.08 & 11.44 & \textbf{36.01} & 46.36 & \textbf{15.03} \\
    CJK & 23.95 & 29.18 & 11.60 & 35.64 & 46.91 & 14.71 \\
    Dravidian & 24.21 & 29.17 & 11.69 & 35.51 & 46.66 & 14.45 \\
    \midrule
    Random (same init) ($1:1$) & 24.21 & 28.89 & 11.86 & 36.07 & 46.03 & 14.79 \\
    Random (same init) ($1:3$) & 24.10 & 29.65 & 11.55 & 37.10 & 46.66 & 15.39 \\
    \midrule
    Only random (same init) & - & - & - & 36.84 & 45.01 & 14.90 \\
    \bottomrule
    \end{tabular}}
    \caption{
    Mean BLEU scores on Flores-200 (after 1 run).
    Results follow adapter training defaults used in the \citet{ustun-etal-2021-multilingual} paper.
    Encoder and decoder adapters are trained together and trained further using cross-attention fine-tuning (CA-FT), and 
    helper and target language adapters are averaged (souped) together.
    \textbf{These scores represent preliminary results, where significantly less data (800 segments) were used to do CA-FT, and thus these results do not correspond to Table~\ref{tab:results}.}
    }
    \label{tab:tgt_avg_flores_old}
\end{table*}

\section{Training Loss}\label{app:loss}

\begin{figure*}
    \centering
    \includegraphics[width=\linewidth]{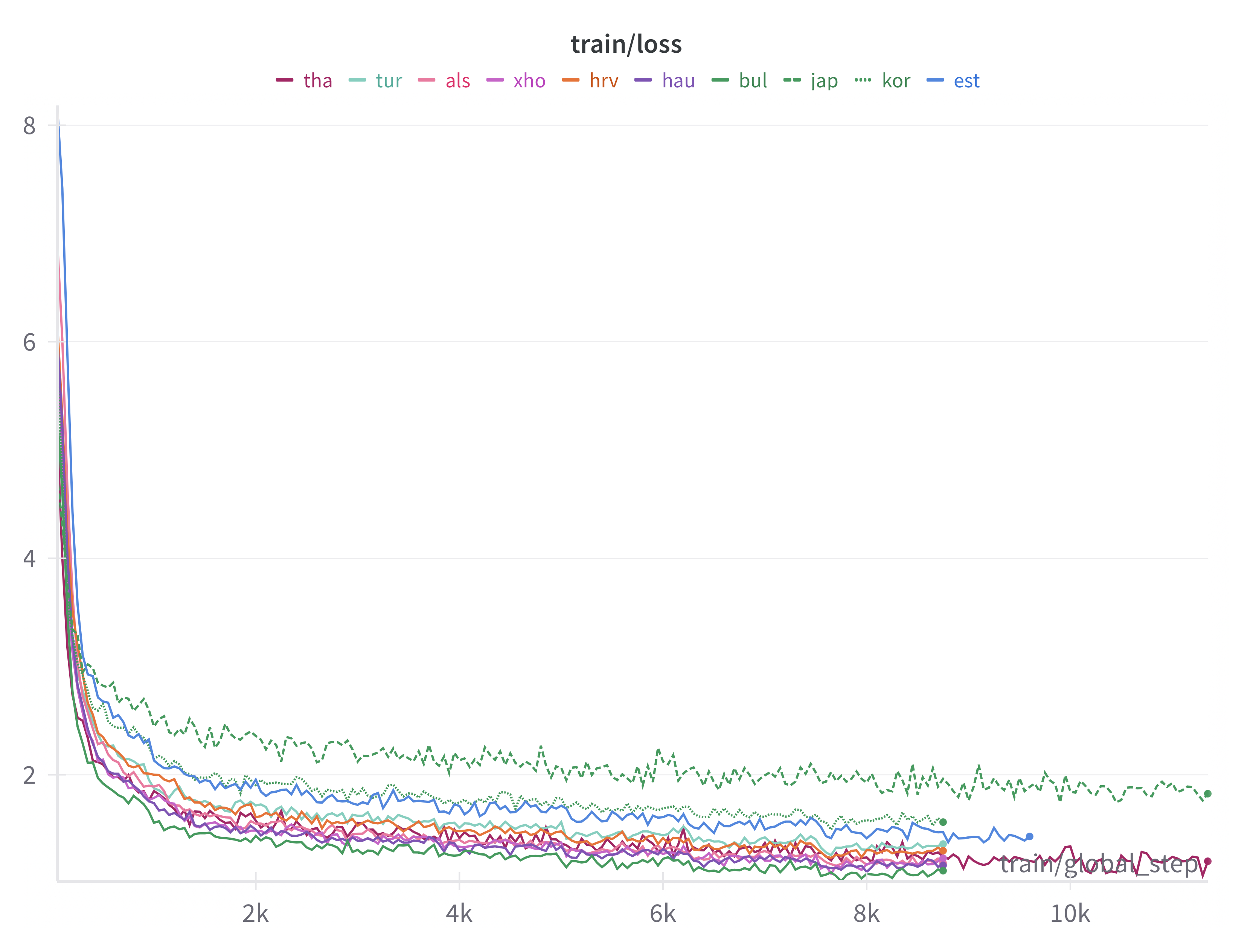}
    \caption{
    Training loss curves for a subset of adapter training experiments.
    }
    \label{fig:loss_curves}
\end{figure*}

Table \ref{tab:losses} compares loss when evaluating on a small subset of the NLLB CA-FT training data.
Figure \ref{fig:loss_curves} indicates that the adapters we train learn successfully from the MADLAD-400 data on a selection from all language adapters.

\begin{table*}[]
    \centering
    \begin{tabular}{lrrr}
    \toprule
    Experiment & hat $\rightarrow$ eng & pap $\rightarrow$ eng & sag $\rightarrow$ eng \\
    \midrule
    $s$ and $t$ Adapters & 0.594 & 1.106 & 1.908 \\
    Untrained $s$ Adapter & \textbf{0.476} & \textbf{0.917} & \textbf{1.796} \\
    \bottomrule
    \end{tabular}
    \caption{
    Validation loss when evaluating on the CA-FT training data.
    Averaging with the weights with the same initialization results in lower loss values.
    }
    \label{tab:losses}
\end{table*}

\begin{table*}[]
    \centering
    \begin{tabular}{lrrrr}
    \toprule
    Experiments & en $\rightarrow$ hu & en $\rightarrow$ yo & hu $\rightarrow$ en & yo $\rightarrow$ en \\
    \midrule
    dNLLB-200 & 22.59 & 8.85 & 29.51 & 18.82 \\
    \midrule
    \texttt{lang2vec} & 21.85 & \textbf{8.51} & 33.38 & 19.27 \\
    NLLB & 21.75 & 8.45 & 33.31 & 19.14 \\
    SuE & \textbf{21.90} & 8.37 & 33.47 & 19.56 \\
    Random (init) ($1:3$) & 21.88 & 8.47 & \textbf{33.97} & \textbf{20.70} \\
    \midrule
    Difference & \cellcolor{orange} $-0.69$ & \cellcolor{orange} $-0.34$ & \cellcolor{green} $+4.46$ & \cellcolor{green} $+1.88$ \\ 
    \bottomrule
    \end{tabular}
    \caption{Mean BLEU scores with our methodology, as applied to Yoruba and Hungarian. \textbf{These scores represent preliminary results, where significantly less data (800 segments) were used to do CA-FT, and thus not directly comparable to Table~\ref{tab:results}.}}
    \label{tab:my_label}
\end{table*}

\section{Native Speaker Evaluation}\label{app:mqm_errors}

See Table \ref{tab:mqm_errors} for a condensed categorization of translation errors under the Untrained Souping (\includegraphics[scale=0.02]{emoji/radio.png}), the IE Transfer (\includegraphics[scale=0.02]{emoji/tree.png}), and the $s$ and $t$ Adapters ($s + t$) conditions.

\begin{table*}[]
    \centering
    \resizebox{\textwidth}{!}{
    \begin{tabular}{lcccl}
    \toprule
    Error & \includegraphics[scale=0.02]{emoji/radio.png} & \includegraphics[scale=0.02]{emoji/tree.png} & $s + t$ & Notes \\
    \midrule
    \midrule
    \multicolumn{5}{c}{\textbf{Grammar Issues}} \\
    \midrule
    \midrule
    \textit{yo rann yo} & \xmark & \xmark & \xmark & Aspectual issue: English translation is passive, Creole is a voluntary activity \\
    \textit{te eksplore} & \xmark & \xmark & \xmark & Aspectual error: English passive which also affects the sentence meaning \\
    "many beautiful buildings to look at" & \cmark & \xmark & \xmark & \includegraphics[scale=0.02]{emoji/tree.png} and $s+t$ ungrammatical \\
    "improve a considerable amount" & \cmark & \xmark & \xmark &  \includegraphics[scale=0.02]{emoji/tree.png} and $s+t$ ungrammatical \\
    "often provide some" & \cmark & \xmark & \cmark & \includegraphics[scale=0.02]{emoji/tree.png} ungrammatical \\
    \textit{fatige mwens posib} & \xmark & \xmark & \xmark & \includegraphics[scale=0.02]{emoji/radio.png} omits it; \includegraphics[scale=0.02]{emoji/tree.png} and $s+t$ to "fatigued less possibly" \\
    \textit{konesans yo bay la} & \xmark & \xmark & \xmark & to "they give/provide" with no clear subject (passive would be better) \\
    \textit{manny\`e efikas} & \xmark & \xmark & \xmark & missing preposition "in" ("in efficient manner") \\
    \midrule
    \midrule
    \multicolumn{5}{c}{\textbf{Translation Issues}} \\
    \midrule
    \midrule
    \multicolumn{5}{l}{Lexical errors} \\
    \midrule
    \textit{nimewotation} & \xmark & \xmark & \xmark & $s+t$ wrongly translates it to "naming system" \\
    \textit{benevola} & \xmark & \xmark & \xmark & "travel benevola" or "benevola travel" (in fact "charity") \\
    \textit{charabya} & \xmark & \xmark & \xmark & Means "nonsense", "gibberish" \\
    \textit{linets} & \xmark & \xmark & \xmark & \includegraphics[scale=0.02]{emoji/radio.png} to "lenses"; \includegraphics[scale=0.02]{emoji/tree.png} to "links"; $s+t$ not translating \\
    \textit{dy\`el} & \xmark & \cmark & \xmark & \includegraphics[scale=0.02]{emoji/radio.png} to "deals"; \includegraphics[scale=0.02]{emoji/tree.png} correctly; $s+t$ to "dyed" \\
    \textit{patikilarite} & \cmark & \cmark & \xmark & $s+t$ to "patikilarity" \\
    \textit{kotidyen} & \cmark & \cmark & \xmark & $s+t$ not translated \\
    \textit{s\`etal\`o} & \xmark & \xmark & \xmark & \includegraphics[scale=0.02]{emoji/radio.png} to "sevental"; \includegraphics[scale=0.02]{emoji/tree.png} to "most"; $s+t$ to "satellite" (in fact "sometimes") \\
    \textit{bildinn} & \xmark & \xmark & \xmark & \includegraphics[scale=0.02]{emoji/radio.png} to "image"; \includegraphics[scale=0.02]{emoji/tree.png} to "figurine"; $s+t$ to "picture" (in fact "building") \\
    \textit{batiman} & \xmark & \xmark & \xmark & \includegraphics[scale=0.02]{emoji/radio.png} and \includegraphics[scale=0.02]{emoji/tree.png} to "vessels"; $s+t$ to "baths" (in fact "building") \\
    \textit{manny\`e} & \cmark & \xmark & \xmark & \includegraphics[scale=0.02]{emoji/radio.png} correctly to "manner"; \includegraphics[scale=0.02]{emoji/tree.png} and $s+t$ to "manny" \\
    \textit{touch} & \xmark & \xmark & \xmark & translated to "touch" though it means "(piano) key" \\
    \textit{foumi milit\`e} & \xmark & \xmark & \xmark & \includegraphics[scale=0.02]{emoji/radio.png} and \includegraphics[scale=0.02]{emoji/tree.png} to "military foam"; $s+t$ to "military fountains" (in fact "army ants") \\
    \textit{py\`ej} & \xmark & \xmark & \xmark & \includegraphics[scale=0.02]{emoji/radio.png} to "hostage"; \includegraphics[scale=0.02]{emoji/tree.png} and $s+t$ to "pigeon" (in fact "trap") \\
    \textit{jako} & \xmark & \xmark & \xmark & \includegraphics[scale=0.02]{emoji/radio.png} and \includegraphics[scale=0.02]{emoji/tree.png} to "jaco"; $s+t$ omitted/misinterpreted (in fact "parrot") \\
    \textit{eskrim\`e} & \xmark & \xmark & \xmark & \includegraphics[scale=0.02]{emoji/radio.png} and $s+t$ to "skirmish"; \includegraphics[scale=0.02]{emoji/tree.png} to "skater" (in fact "fencer") \\
    \textit{parese} & \xmark & \xmark & \xmark & to "parasites" (in fact "sloths") \\
    \textit{pwomennen} & \xmark & \xmark & \xmark & to "promoted" (in fact "walking") \\
    \textit{eta espri} & \cmark & \cmark & \xmark & \includegraphics[scale=0.02]{emoji/radio.png} and \includegraphics[scale=0.02]{emoji/tree.png} correctly to "state of mind"; $s+t$ to "spirit state" \\
    \textit{Premye/Twazy\`em Group} & \xmark & \xmark & \xmark & to "First/Third Group" (in fact "First/Third Class") \\
    \textit{mache} & \xmark & \xmark & \xmark & to "walk" (in fact "migrate" in context) \\
    \midrule
    \multicolumn{5}{l}{Other errors} \\
    \midrule
    \textit{kote pasaj enp\'otan yo} & \xmark & \xmark & \xmark & \includegraphics[scale=0.02]{emoji/radio.png} and \includegraphics[scale=0.02]{emoji/tree.png} too literal; $s+t$ also adding "their" instead of "the" \\
    \bottomrule
    \end{tabular}
    }
    \caption{Condensed categorization errors under various translation conditions.
    \includegraphics[scale=0.02]{emoji/radio.png} refers to Untrained Souping, \includegraphics[scale=0.02]{emoji/tree.png} indicates IE Transfer, and $s+t$ means $s$ and $t$ Adapters.
    }
    \label{tab:mqm_errors}
\end{table*}

\end{document}